\label{key}
\documentclass[letterpaper, 10 pt, journal, twoside]{IEEEtran}
\usepackage{amsmath,amsfonts}	% assumes amsmath package installed
\usepackage{amssymb}	% assumes amsmath package installed
\usepackage[linesnumbered,ruled]{algorithm2e}
\usepackage{array}
\usepackage[caption=false,font=footnotesize,labelfont=rm,textfont=rm]{subfig}
\usepackage{textcomp}
\usepackage{stfloats}
\usepackage{url}
\usepackage{verbatim}
\usepackage{graphicx}
\usepackage{cite}
\usepackage{booktabs}
\usepackage{makecell}
\usepackage{multirow}

\usepackage{hyperref} % 超链接
\begin{document}

\title{Aligning Human Intent from Imperfect Demonstrations with Confidence-based \\Inverse soft-Q Learning}

\author{Xizhou Bu, Wenjuan Li, Zhengxiong Liu, Zhiqiang Ma, and Panfeng Huang
        % <-this % stops a space
\thanks{Manuscript received: February, 22, 2024; Revised April, 8, 2024; Accepted June, 6, 2024. This paper was recommended for publication by Editor Sven Behnke upon evaluation of the Associate Editor and Reviewers' comments. This work was supported in part by the National Natural Science Foundation of China under Grant 62273280 and 62373305, and in part by the Aerospace Flight Dynamics Technology Key Laboratory Foundation of China under Grant KJW6142210210303. \textit{(Corresponding author: Panfeng Huang.)}}
\thanks{Xizhou Bu, Wenjuan Li, Zhengxiong Liu, Zhiqiang Ma, and Panfeng Huang are with the School of Astronautics, Northwestern Polytechnical University, Xi'an, Shaanxi 710072 China. (e-mail: piaoxi@mail.nwpu.edu.cn; wenjuanli@mail.nwpu.edu.cn; liuzhengxiong@nwpu.edu.cn; zhiqiangma@ nwpu.edu.cn; pfhuang@nwpu.edu.cn)}
\thanks{$^{1}$ Our implementation is available at \href{https://github.com/XizoB/CIQL}{https://github.com/XizoB/CIQL}}
}

\markboth{IEEE Robotics and Automation Letters. Preprint Version. Accepted June, 2024}%
{Bu \MakeLowercase{\textit{et al.}}: Aligning Human Intent from Imperfect Demonstrations with Confidence-based Inverse soft-Q Learning}


\maketitle

\begin{abstract}
Imitation learning attracts much attention for its ability to allow robots to quickly learn human manipulation skills through demonstrations. However, in the real world, human demonstrations often exhibit random behavior that is not intended by humans. Collecting high-quality human datasets is both challenging and expensive. Consequently, robots need to have the ability to learn behavioral policies that align with human intent from imperfect demonstrations. Previous work uses confidence scores to extract useful information from imperfect demonstrations, which relies on access to ground truth rewards or active human supervision. In this paper, we propose a transition-based method to obtain fine-grained confidence scores for data without the above efforts, which can increase the success rate of the baseline algorithm by 40.3$\%$ on average. We develop a generalized confidence-based imitation learning framework for guiding policy learning, called Confidence-based Inverse soft-Q Learning (CIQL), as shown in Fig.\ref{algo}. Based on this, we analyze two ways of processing noise and find that penalization is more aligned with human intent than filtering.\hyperref[opensoure]{$^1$}
\end{abstract}


\section{Introduction}
\IEEEPARstart{I}{Imitation} Learning (IL) typically assumes that demonstrations are drawn from an optimal policy distribution. It depends on large, high-quality and diverse datasets to achieve satisfactory results. In the real world, collecting such human datasets is very challenging and costly\cite{pinto2016supersizing,mandlekar2021matters,beliaev2022imitation}. Due to fatigue, distraction, or lack of skills, the human datasets may contain various types of noise, including random behavior, bias, or unnecessary operations\cite{mandlekar2019scaling}. These noises can adversely affect the performance of standard imitation learning\cite{tangkaratt2019vild}. While demonstrating perfect trajectories is a challenge for humans, poorly performed or unsuccessful trajectories are not without value, as they may contain some genuine human intents. Overall, it is difficult for robots to accurately align with the true human intent from these imperfect demonstrations. This paper aims to maximize the use of limited imperfect demonstrations to learn a policy aligned with human intent.

Learning from imperfect demonstrations has been developed for a long time, and its work mainly falls into two directions. Firstly, The ranking-based methodology learns a reward function satisfying the ranking of trajectories through supervised learning, and subsequently improves the policy through reinforcement learning\cite{brown2019extrapolating,brown2020better,chen2021learning}. It aims to train policies that outperform the demonstrator's performance rather than aligning human intent. In addition, it requires high-quality datasets to learn a robust reward function. On the other hand, the confidence-based methodology employs confidence scores as a measure of the demonstration's quality so that useful information can be extracted to guide policy learning, and the challenge is to obtain the ground truth confidence scores for the data\cite{wang2021learning,wu2019imitation,zhang2021confidence}.

\begin{figure}[t]
	\centering
	\includegraphics[width=0.47\textwidth]{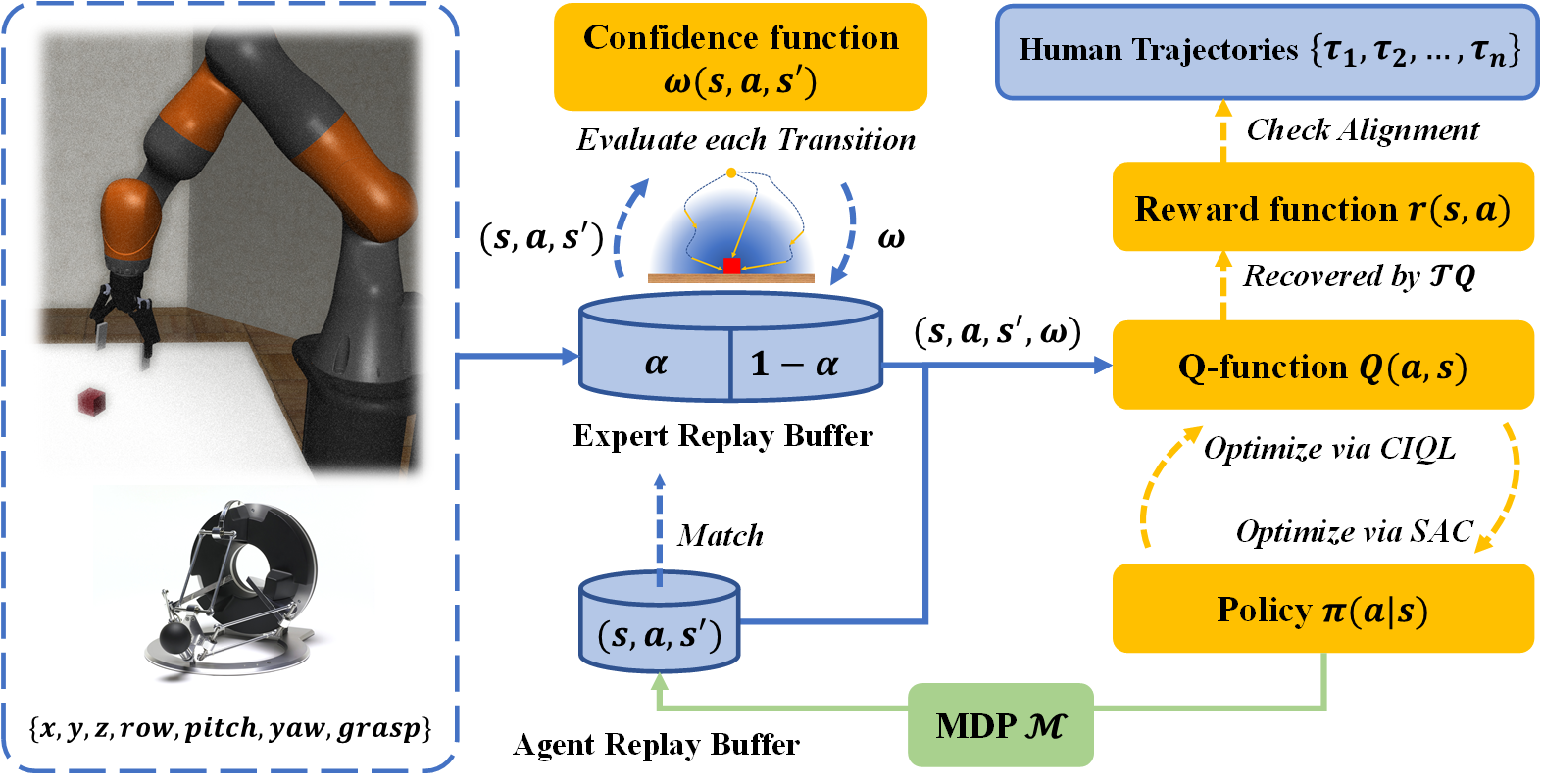}
	\caption{\textbf{Overview of Confidence-based Inverse soft-Q Learning:} 
		1) Evaluate each transition $(s,a,s')$ using the confidence function $w$ and estimate the optimal prior probability $\alpha$.
		2) Look for the optimal Q-function by CIQL and search for the optimal policy by Soft Actor-Critic (SAC).
		3) Recover the reward function $r(s,a)$ using the inverse soft Bellman operator $\mathcal{T}^{\pi}$ and evaluate whether it aligns human intent.}
	\label{algo}
\end{figure} 

Assessing every data point within a trajectory can be impractical for humans, while evaluating an entire trajectory or segment may result in overly coarse estimates. To obtain fine-grained confidence scores for trajectories, several methods are commonly employed:
1) Assuming that human experts exhibit stable performance or making strict assumptions about noise \cite{beliaev2022imitation,cao2021learning, tangkaratt2019vild} can be problematic due to factors like human fatigue or distraction.
2) Training a classifier on partially labeled data to predict confidence scores for unlabeled data may lead to error accumulation and performance degradation\cite{wu2019imitation}.
3) Using reward functions recovered through inverse reinforcement learning to evaluate data\cite{zhang2021confidence,wang2021learning,cao2021learning} may produce inaccurate results and adversely affect performance due to the inherent ambiguity of reward functions\cite{kingma2014semi}. In summary, these methods often rely on strict assumptions, active human supervision, or ground truth rewards and do not provide accurate confidence scores.

To mitigate the above problems, we propose a transition-based confidence evaluation method and introduce the noise angle to provide criterion for the evaluation. This method allows to obtain a confidence score for each transition by simply setting the noise angle. In short, it's a way to determine if a robot is approaching to the target in a straight line. Through experiments, we find the optimal range of values for the noise angle. Based on state-of-the-art imitation learning, we integrate a generalized confidence-based imitation learning framework called \textit{Confidence-based Inverse soft-Q Learning} to validate our method and provide two ways to learn the optimal policy. Our main contributions are as follows: 

1) We propose a transition-based confidence evaluation method to obtain fine-grained confidence scores for demonstrations without any cost. 

2) We integrate a generalized confidence-based imitation learning framework, provide two methods for learning optimal policy, and analyze their differences in processing noise. 

3) We experimentally find that the optimal noise angle for linear task ranges from 20° to 60°, as well as that penalizing noise is more aligned with human intent than filtering noise.

\section{Related Work}
Recall that our goal is to maximize the use of a limited number of imperfect human demonstrations to learn a policy that is aligned with human intent. In other words, our goal is twofold: one, we desire to be able to utilize all data including noise in the demonstrations; and two, we want to learn a policy or a reward function that is aligned with human intent. Therefore, we focus on three main things: how to evaluate this data costlessly, how to use these confidence scores for imitation learning, in which we focus more on how to process noise, and how to do alignment testing.

\textbf{Imitation Learning:} 
Behavioral Cloning (BC) is a simple imitation learning that directly minimizes the difference in action probability distribution between expert and learned policy via supervised learning\cite{bain1995framework}. It suffers from the problem of compounding error where small errors in imitation learning will gradually amplify as the decision sequence continues\cite{ross2010efficient}. Inverse Reinforcement Learning (IRL) recovers an expert's policy by inferring the expert's reward function and then finding the policy that maximizes rewards under this reward function\cite{abbeel2004apprenticeship}. This framing can use the environmental dynamics to reduce compounding error, which has inspired several methods\cite{xu2020error}. Based on this, Generative Adversarial Imitation Learning (GAIL) learns a policy by matching the occupancy measure between expert and learned policy in an adversarial manner\cite{ho2016generative}. In this process, GAIL's discriminator provides an implicit reward function to separate different policies. Adversarial Inverse Reinforcement Learning (AIRL) recovers an explicit reward function on GAIL's discriminator that is robust to changes in the environmental dynamics, leading to more robust policy learning\cite{fu2017learning}. These methods require modeling reward and policy separately and training them in an adversarial manner, which is difficult to train in practice\cite{baram2017end}. Recently, Inverse soft-Q Learning (IQ-Learn) has achieved state-of-the-art results, which avoids adversarial training by learning a single Q-function that implicitly represents both reward and policy\cite{garg2021iq}.

\textbf{Confidence-based Imitation Learning:}
The confidence-based methodology assumes that each demonstration or demonstrator has a confidence score that indicates its optimality, to reweight the demonstrations based on that score for imitation learning. There are two main methods for using data weighted with confidence scores to guide policy learning. One is to directly estimate the distribution of optimal policy in the demonstrations\cite{wu2019imitation,wang2021learning,zhang2021confidence,cao2021learning,beliaev2022imitation,tangkaratt2019vild}. This method is preferred by current researches because it is simple and easy to implement. The other is to estimate the distribution of non-optimal policies and learn the optimal policy based on the mixed distribution setting\cite{wu2019imitation}. We find that these two methods process noise in different ways and we analyze their impact on the performance of the baseline algorithm. Moreover, different from previous work which are all based on GAIL, AIRL or related variants, our framework uses the state-of-the-art imitation learning, IQ-Learn, as the baseline algorithm.

\textbf{Human Intent Alignment:}
As the application domain of deep reinforcement learning continues to expand, concerns have been raised about the potential for agents to exhibit adverse behaviors. Reflections on the safety of agents and the need to deploy agents in real-world environments inspire research on alignment theory. Alignment aims to ensure that the behavior of an agent conforms to the human intent, which often exhibits diversity in the form of instructions, desires, and preferences\cite{gabriel2020artificial}. Existing methods primarily focus on understanding the causes of alignment failures, such as reward manipulation\cite{leike2018scalable} and errors in generalizing goals\cite{di2022goal}, as well as studying alignment concepts and mathematical models. Ji et al.\cite{ji2023ai} classify alignment into two categories: alignment training and alignment refinement. The former involves learning from feedback\cite{brown2019extrapolating,zhang2021confidence} or adapting to distributional shifts\cite{bai2021recent,dafoe2020open}, while the latter ensures alignment through evaluation and assurance\cite{anderljung2023frontier, shevlane2023model}. It is worth mentioning that Bobu et al.\cite{bobu2022inducing} propose a general feature-based reward learning framework called FERL, which enables reward learning from various types of human inputs. Inspired by FERL's visualization of reward functions in the robot's task space, we employ learned reward function to evaluate human demonstrations and determine whether it align with human intent.

\section{Methods}
In this section, we describe our framework in detail, including two confidence-based Inverse soft-Q Learning objectives, a transition-based confidence evaluation method to label transitions with fine-grained confidence scores, and specific steps for algorithm implementation.
\subsection{Problem Setting}
We consider the standard Markov Decision Process (MDP) to represent the robot's sequential decision-making task, which is defined as the tuple $\mathcal{M}=(\mathcal{S},\mathcal{A},\mathcal{P},\rho_0,r,\gamma)$. $\mathcal{S,A}$ are state and action spaces, $\mathcal{P}:\mathcal{S}\times\mathcal{A}\times\mathcal{S}\rightarrow[0,1]$ is the transition probability, as known as dynamics, $\rho_0$ is the initial state distribution, $r:\mathcal{S}\times\mathcal{A}\rightarrow\mathbb{R}$ is the reward function, and $\gamma\in(0,1)$ is the discount factor. $\mathbb{R}^{\mathcal{S}\times\mathcal{A}}=\{x:\mathcal{S}\times\mathcal{A}\rightarrow\mathbb{R}\}$ denotes the set of all functions in the state-action space.

A policy $\pi:\mathcal{S}\times\mathcal{A}\rightarrow[0,1]$ is defined as a probability distribution over the space of actions in a given state. For a policy $\pi$, we define its occupancy measure $\rho_{\pi}:\mathcal{S}\times\mathcal{A}\rightarrow\mathbb{R}$ as $\rho_{\pi}(s,a)=(1-\gamma)\pi(a|s)\Sigma^{\infty}_{t=0}\gamma^tP(s_t=s|\pi)$, where $P(s_t=s|\pi)$ is the probability distribution of state $s$ at step $t$ following policy $\pi$, and $\rho_{\pi}(s,a)$ can be interpreted as a normalized distribution of state-action pairs. The occupancy measure is important because it has a one-to-one correspondence with the policy\cite{puterman2014markov}.  We refer to the expert policy as $\pi_E$ and its occupancy measure as $\rho_E$. Reinforcement learning aims to learn an optimal policy $\pi^*$ that maximizes the expected cumulative discounted reward, $\mathbb{E}_{s_0\sim\rho_0,a\sim\pi(\cdot|s)}[\sum^{\infty}_{t=0}\gamma^tr(s_t,a_t)]$, which can equivalently be expressed in terms of the occupancy measure, $\mathbb{E}_{\rho_{\pi}}[r(s,a)]$.
\subsection{Background}
\textbf{Inverse Reinforcement Learning:} 
It aims to recover the reward function that can rationally explain the expert's behavior by solving the optimization problem\cite{ho2016generative}, $\mathop{\max}_{r}\mathop{\min}_{\pi}L(\pi,r)=\mathbb{E}_{\rho_E}[r(s,a)]-\mathbb{E}_{\rho_{\pi}}[r(s,a)]-H(\pi)$, where $H(\pi)=\mathbb{E}_{\rho_{\pi}}[-log\ \pi(a|s)]$ is the $\gamma$-discounted causal entropy of the policy $\pi$ to encourage exploration. The IRL objective optimization process is to find a reward function that assigns high reward to the expert policy and low reward to other ones, while searching for the best policy for the reward function in an inner loop. In general, we can exchange the max-min resulting in an objective that minimizes the statistical distance between the expert and the policy\cite{ho2016generative}:
\begin{equation}
	\mathop{\min}_{\pi}\mathop{\max}_{r}L(\pi,r)=\mathop{\min}_{\pi}d(\rho_{\pi},\rho_{E})-H(\pi)
\end{equation}
where $d$ is a distance function that can be expressed as:
\begin{equation}
	d(\rho_{\pi},\rho_{E})=\mathop{\max}_{r}\mathbb{E}_{\rho_E}[\phi(r(s,a))]-\mathbb{E}_{\rho_{\pi}}[r(s,a)]
\end{equation}
where $\phi$ is a statistical distance that allows for Integral Probability Metrics (IPMs) and f-divergences. In this paper, for the choice of $\phi$, we use $\mathcal{X}^2$-divergence, specifically, $\phi(x)=x-\frac{1}{4\sigma}x^2$ with $\sigma=0.5$. The structure of the divergence reward function induced by it is\cite{garg2021iq}:
\begin{align}
	r(s,a)=2(1-\rho_{\pi}/\rho_E)
	\label{divreward}
\end{align}
which is done to more efficiently analyze the contribution of the expert data distribution to the learned reward function, as detailed in Section \ref{CIQL}.

\textbf{Inverse soft-Q Learning:}
It develops from IRL and introduces an inverse soft Bellman operator that transforms the reward function into a Q-function to avoid learning the reward function directly. The definition of the inverse soft Bellman operator $\mathcal{T}^{\pi}:\mathbb{R}^{\mathcal{S}\times\mathcal{A}\rightarrow\mathcal{S}\times\mathcal{A}}$ is given as\cite{garg2021iq}:
\begin{equation}
	(\mathcal{T}^{\pi}Q)(s,a)\triangleq Q(s,a)-\gamma\mathbb{E}_{s'\sim \mathcal{P}(\cdot|s,a)}V^{\pi}(s')
	\label{ibo}
\end{equation}
where $V^{\pi}(s)=\mathbb{E }_{a\sim\pi(\cdot|s)}[Q(s,a)-log\ \pi(a|s)]$ is the soft state value function \cite{haarnoja2018soft}. $\mathcal{T}^{\pi}$ is bijective, and for $r=\mathcal{T}^{\pi}Q$, we can freely transform between $Q$ and $r$\cite{garg2021iq}. In other words, we can obtain the reward function for any Q-function by repeatedly applying $\mathcal{T}^{\pi}$. For brevity, we do not expand the exact form of the definition of $\mathcal{T}^{\pi}Q$ in the following sections. Using the $r=\mathcal{T}^{\pi}Q$ equivalently transforms the original IRL objective into the IQ-Learn's objective:
\begin{equation}
	\mathop{\max}_{Q}\mathop{\min}_{\pi}J(\pi,Q)=\mathbb{E}_{\rho_E}[\phi(r)]-\mathbb{E}_{\rho_{\pi}}[r]-H(\pi)
\end{equation}

Noting that although IQ-Learn's objective does not directly recover the reward function during training, it can still do IRL by recovering the reward function via $r=\mathcal{T}^{\pi}Q$. Thus, IQ-Learn can be used as a baseline algorithm for different confidence evaluation methods, even those require real-time recovery of the reward function\cite{wang2021learning,zhang2021confidence,cao2021learning}.
\subsection{Confidence-based IQ-Learn}
\label{CIQL}
We assume that the human imperfect demonstrations are sampled from an optimal policy $\pi_{E_{opt}}$ and the non-optimal policies $\pi_{E_{non}}=\{\pi_{E_i}\}^n_{i=1}$. Following the method presented in paper\cite{wu2019imitation}, we define $\rho_{opt}(s,a)=\rho_E(s,a|y=+1)$ and $\rho_{non}(s,a)=\rho_E(s,a|y=-1)$, where $y=+1$ indicates that the state-action pair $(s,a)$ is drawn from the optimal policy, and $y=-1$ indicates that it is drawn from the non-optimal policies. Then we define the confidence scores of each state-action pair as $w(s,a)=\rho_E(y=+1|s,a)$. Moreover, $\alpha=p(y=+1)$ denotes the priori probability of the optimal policy, which can be estimate by $\alpha\approx\frac{1}{N}\sum_{i=1}^{N}w_i$, where $N$ is the total number of data in demonstrations. Based on this, the normalized occupancy measure of the optimal policy can by expressed by the Bayes' rule as:
\begin{equation}
	\rho_{opt}(s,a) = \frac{w(s,a)}{\alpha}\rho_E(s,a)
	\label{optimal}
\end{equation}

Correspondingly, the normalized occupancy measure of the non-optimal policies can be expressed as:
\begin{equation}
	\rho_{non}(s,a) = \frac{1-w(s,a)}{1-\alpha} \rho_E(s,a)
	\label{nonoptimal}
\end{equation}

Therefore, the occupancy measure of the expert policy can be expressed as follows:
\begin{equation}
	\rho_E(s,a)=\alpha\rho_{opt}(s,a)+(1-\alpha)\rho_{non}(s,a)
	\label{expertdistribution}
\end{equation}

The main idea of confidence-based imitation learning is $\mathop{\min}_{\pi}d(\rho_{\pi},\rho_{opt})$, which matches $\rho_{opt}$ by minimizing the statistical distance between the expert's optimal policy and the learned policy. In practice, the expert policy $\pi_E$ and its occupancy meansure $\rho_E$ are unknown and we have access to a sampled datasets of demonstrations. If we can obtain confidence scores for all the data in demonstrations, we can estimate the values of $\rho_{opt}$ and $\rho_{non}$. Based on this, there are two methods that can guide confidence-based policy learning.

\textbf{CIQL-E (Expert):} 
The first method is to learn the policy by directly matching the occupancy measure of the optimal policy, $\rho_{opt}$, which can be estimated by Eq.(\ref{optimal}). Through $\mathop{\min}_{\pi}d(\rho_{\pi},\rho_{opt})\Rightarrow d(\rho_{\pi},\frac{w}{\alpha}\rho_E)$ we can transform the IQ-Learn's objective into CIQL-E's objective as:
\begin{equation}
	\begin{aligned}
		\mathop{\max}_{Q}\mathop{\min}_{\pi}J(\pi,Q)=\ 
		\mathbb{E}_{\rho_E}[\frac{w}{\alpha}\cdot\phi(r)]-\mathbb{E}_{\rho_{\pi}}[r]-H(\pi)
		\label{optimalIQ}
	\end{aligned}
\end{equation}
which only makes assumptions about the distribution of expert demonstrations. According to Eq.(\ref{divreward}), its divergence reward function can be expressed as:
\begin{equation}
	\begin{aligned}
		r(s,a)=2(\frac{\rho_{opt}-\rho_{\pi}}{\rho_{opt}})=2(\frac{w\rho_{E}-\alpha\rho_{\pi}}{w\rho_{E}})
	\end{aligned}
	\label{divrewardciqle}
\end{equation}

\textbf{CIQL-A (Agent):} 
The second method assumes that the learned policy distribution structure is the same as that of the expert, $\rho_{\pi'} = \alpha\rho_{\pi} + (1-\alpha)\rho_{non}$. Recalling Eq.(\ref{expertdistribution}), if the divergence between $\rho_{\pi'}$ and $\rho_{E}$ is minimized, then the divergence between $\rho_{\pi}$ and $\rho_{opt}$ will be minimized as well. Eventually through $\mathop{\min}_{\pi}d(\rho_{\pi'},\rho_{E})\Rightarrow d(\alpha\rho_{\pi}+(1-\alpha)\rho_{non},\rho_E)\Rightarrow d(\alpha\rho_{\pi}+(1-w)\rho_{E},\rho_E)$ we can transform IQ-Learn's objective into CIQL-A's objective as:
\begin{equation}
	\begin{aligned}
		\mathop{\max}_{Q}\mathop{\min}_{\pi}J(\pi,Q)=\ &
		\mathbb{E}_{\rho_E}[\phi(\mathcal{T}^{\pi}Q)-(1-w)\cdot\mathcal{T}^{\pi}Q]
		\\
		&-\alpha\mathbb{E}_{\rho_{\pi}}[\mathcal{T}^{\pi}Q]-H(\pi)
	\end{aligned}
	\label{nonoptimalIQ}
\end{equation}
where $\rho_{non}$ can be estimated by Eq.(\ref{nonoptimal}). According to Eq.(\ref{divreward}), The structure of its divergence reward function can be expressed as:
\begin{equation}
	\begin{aligned}
		r(s,a)=2(\frac{\rho_{E}-\rho_{\pi'}}{\rho_{E}})=2(\frac{w\rho_{E}-\alpha\rho_{\pi}}{\rho_{E}})
	\end{aligned}
\end{equation}
It is noted that CIQL-A brings an additional benefit of being able to utilize the full information from all the data in demonstrations without labeling them\cite{wu2019imitation}. In this work, we have access to the data fine-grained confidence scores without any cost, so we will not take this advantage to ensure the accuracy of our estimates. Algorithm \ref{algorithm} presents the details of our framework, where the policy improvement uses the objective function of updating the Actor in SAC.

\begin{algorithm}[t]
	\caption{Confidence-based IQ-Learn}
	\label{algorithm}
	\KwIn{Human datasets $\mathcal{D}_E$, learn rate $l_{\psi}$, $l_{\varphi}$, ratio $\mu$, noise angle $\theta_n$, Q-function $Q_{\psi}$ and policy $\pi_{\varphi}$}
	
	Calculate the confidence score $w$ for each transition $(s,a,s')$ in $\mathcal{D}_E$ using Eq.(\ref{w}), and estimate the prior probability of the optimal policy by $\alpha\approx\frac{1}{N}\sum_{i=1}^{N}w_i$
	
	\For{$i=0,1,2,...$}{
		Sample the transitions $(s,a,s',w)$ from $\mathcal{D}_E$, $\mathcal{D}_A$
		
		Optimize the Q-function $Q_{\psi}$ using the objective from Eq.(\ref{optimalIQ}) (CIQL-E) or Eq.(\ref{nonoptimalIQ}) (CIQL-A):
		$\psi_{t+1}\leftarrow\psi_{t}+l_{\psi}\cdot\nabla_{\psi}J(\pi,Q_{\psi})$
		
		Improve the policy $\pi_{\varphi}$ by using the Actor's objective in SAC:
		$\varphi_{t+1}\leftarrow\varphi_{t}-l_{\varphi}\cdot\nabla_{\varphi}\mathbb{E}_{s\sim \mathcal{D_A},a\sim\pi_{\varphi}(\cdot|s)}[log\ \pi_{\varphi}(a|s)-Q(s,a)]$
		
		Policy $\pi_{\varphi}$ interacts with the environment $\mathcal{M}$ and add the new transitions to Agent datasets $\mathcal{D}_A$
		
		Jump to step 9 to recover the reward if needed.
	}
	Given a trained Q-function $Q_{\psi}$, and a trained policy $\pi_{\varphi}$, recover the reward function by Eq.(\ref{ibo}):
	$r(s,a)=Q_{\psi}(s,a)-\gamma\mathbb{E}_{s'\sim \mathcal{P}(\cdot|s,a)}V^{\pi_{\varphi}}(s')$.
\end{algorithm}

\textbf{Noise Processing Analysis:} 
Interestingly, we find that these two confidence-based methods induce different divergence reward functions. We analyze their processing of noise by comparing the divergence reward function and the objective. The confidence score corresponding to the noise is $w=0$. When the sampled data is noise, the divergence reward function of CIQL-E is not available through Eq.(\ref{divrewardciqle}). And the first term of the CIQL-E's objective is equal to 0, which indicates that it just filters the noise. That is, the noise does not contribute to the learned reward function. Whereas the divergence reward function of CIQL-A is negative, combined with the fact that the first term in the CIQL-A's objective is also negative, it can be shown that CIQL-A penalizes noise. This can be interpreted that it not only allows the agent to learn how to behave from experts, but also how not to behave from noise. In short, CIQL-E is to filter noise directly, while CIQL-A is to penalize noise.
\subsection{Transition Evaluation}
The highlight of our work is the proposed transition-based confidence evaluation method that can provide fine-grained confidence scores for demonstrations. In brief, it determines whether the next robot's state is towards the target in a straight line. Since the method does not use supervised or self-supervised learning, it does not need to provide any priori information. The disadvantage is that it can only be used for linear robotic tasks. On the other side of the coin, any task with a linear stage can benefit from our method's improvement in the performance of the baseline algorithm.

\textbf{Robot Manipulation Analysis:} 
The diversity of robot tasks with different manipulations makes it necessary to analyze the robot tasks. We simplify the robot manipulation process into two stages, the robot movement stage and the actuator operation stage. 1) In the robot movement stage, our main goal is to find and approach the target, as well as the movements that need to be made in order to accomplish the task, i.e., the robot needs to follow a policy or a human-generated trajectory. For human demonstrations, this stage is particularly prone to errors or noise, especially if the operator is inexperienced. Our method is to evaluate the imperfect data at this stage. 2) In the actuator operation stage, it usually contains key points data, such as grasping, pressing, and other execution actions. This stage typically represents only a small portion of the overall task and may contain only a few data points in the complete trajectory data. For example, the grasping action may just have a simple binary signal that requires one data point. However, these data points are critical to the successful completion of the task. Previous work has focused on manipulation tasks that involve only the robot movement stage\cite{zhang2021confidence,bobu2022inducing,chen2021learning,cao2021learning}. Since the reward function recovered by IRL may miss some potentially important details\cite{castro2019inverse}, such as the grasping data points in the grasping task, which cannot be completed without learning the action of grasping. In summary, our method is used for the robot movement stage, and as for the actuator operation stage, we emphasize the importance of this data by setting the confidence to an interval between 1 and 2.

\begin{figure}[t]
	\centering
	\includegraphics[width=0.48\textwidth]{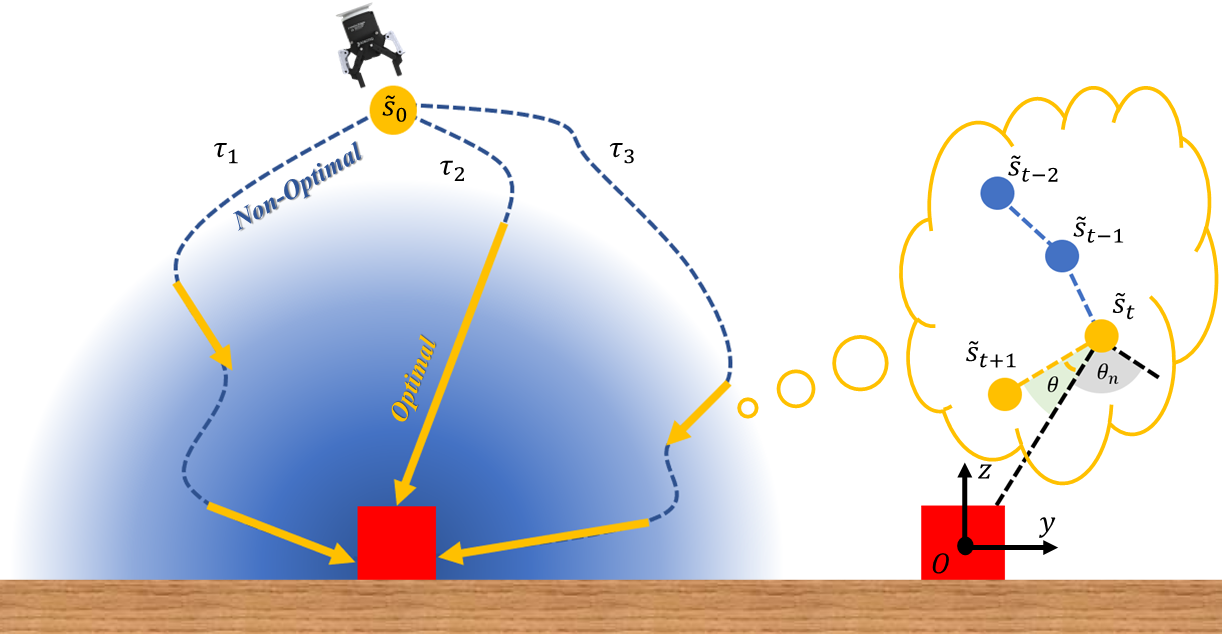}
	\caption{\textbf{Noise Angle Setting.} 
		$\tau_1,\tau_2,\tau_3$ are the complete trajectories. $\tilde{s}_0$ and $o$ are the initial positions of gripper and target, respectively. $\theta$ is the approach angle between vectors ${\tilde{s}_to}$ and ${\tilde{s}_t\tilde{s}_{t+1}}$, and $\theta_{n}$ is the noise angle. We evaluate each transition $(s_t,a_t,s_{t+1})$ using the Eq.(\ref{condition}) as a criterion.}
	\label{Confidence_Evalute}
\end{figure}

\textbf{Linear Task Confidence Setting:} 
We let the demonstrator complete the grasping task quickly, and intent can be expressed in terms of the environmental state reward function, as shown in Fig.\ref{Confidence_Evalute}. The blue gradient circle is the ideal reward function for the operator, where the inner portion is high reward and the outer portion is low reward. To generate this reward function, the yellow trajectory segments close to the target in a straight line are defined as optimal data, while the blue trajectory segments are defined as non-optimal or noise. We introduced the definitions of approach angle $\theta$ and noise angle $\theta_n$, where the approach angle is the angle between the robot's state at this moment to the next state and the target, which is used to determine whether the robot reaches the target in a straight line. We can define it as follows:
\begin{equation}
	\theta(s_t,a_t,s_{t+1})=\arccos(\frac{{\tilde{s}_to}\cdot{\tilde{s}_t\tilde{s}_{t+1}}}{||{\tilde{s}_to}||\cdot||{\tilde{s}_t\tilde{s}_{t+1}}||})
\end{equation}
where the value domain of $\theta$ is $[0, \pi]$. $\tilde{s}_0$ and $o$ are the initial positions of gripper and target, respectively. The data $(\tilde{s}_t,\tilde{s}_{t+1},o)$ are some features in the transition $(s_t, a_t , s_{t+1})$. For a given transition, it has a consistent value $\theta$ that can be used to calculate its confidence score $w$. Therefore, we can categorize the data into three components based on $\theta$: optimal, non-optimal and noise. 
\begin{figure}[t]
	\centering
	\subfloat[Simulated Robotic Teleoperation System]{\includegraphics[width=0.48\textwidth]{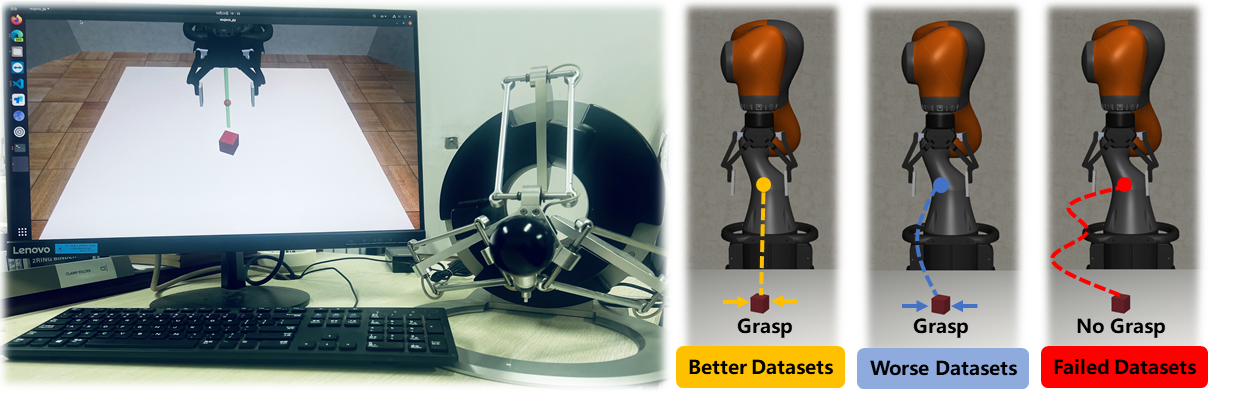}\label{fig:Tele}}
	
	\subfloat[Noise Filtering Visualization]{\includegraphics[width=0.48\textwidth]{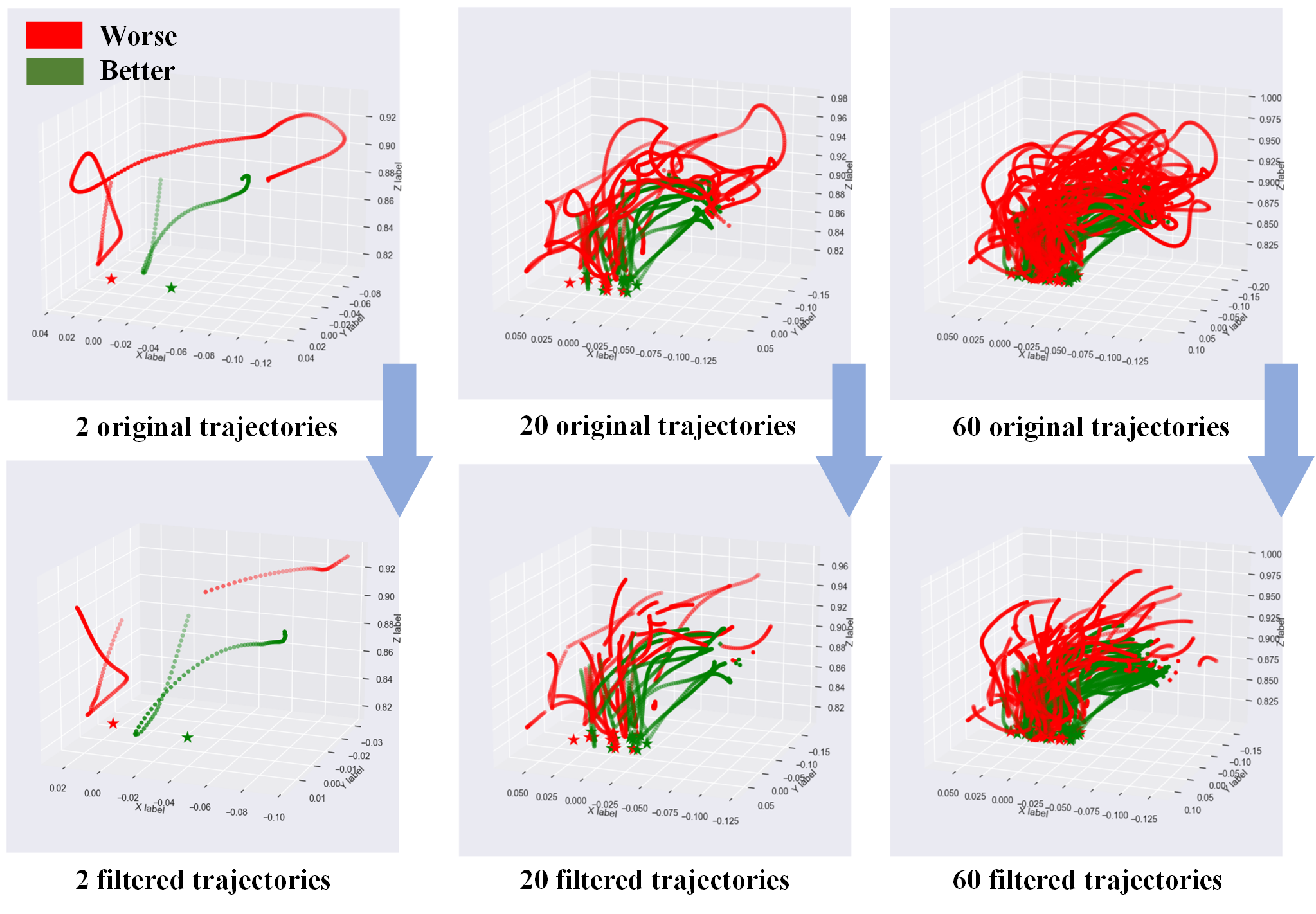}\label{fig:filtering}}
	\caption{\textbf{Demonstration collection and processing.} (a) The system control frequency is 20Hz and the task time limit is 25 seconds, i.e., the length of the trajectory is limited to 500 steps. The intervals of length setting for the different datasets are, Better datasets: 100-150; Worse datasets: 200-400; Failed datasets: 500. (b) Noise filtering visualization of two human datasets, Better and Worse.}
	\label{Dataset}
\end{figure}
\begin{equation}
	w(\theta)=\left\{
	\begin{array}{lll}
		1 &for\quad \theta=0,& opt\\ 
		(0,1) &for\quad 0<\theta\leq \theta_n,& non\ opt\\ 
		0 &for\quad  \theta_n< \theta\leq\pi,& noise
	\end{array}
	\right .
	\label{condition}
\end{equation}
where the noise angle $\theta_n$ is set to provide a classification criterion to determine whether this data is noise. Transitions that satisfy $\theta\leq\theta_n$ are viewed as useful, while others are viewed as noise. In the useful datasets, it can be further subdivided into optimal and non-optimal data. We find a simple function for $w$ that significantly improves the performance of the baseline algorithm:
\begin{equation}
	w(\theta)=\rm Sigmoid((\theta/\theta_n-0.5)*\mu)
	\label{w}
\end{equation}
where the ratio $\mu$ satisfies $\rm Sigmoid(0.5*\mu)=\epsilon$, and $\epsilon$ is the limiting distance from the actual score to the confidence bound. Here, we can assign confidence scores to each transition. Since the dynamics of the robot's environment are deterministic, $w(s_t,a_t,s_{t+1})$ is equivalent to $w(s_t,a_t)$. Note that the confidence function $w$ relies on transition information and does not require other inputs such as ground truth rewards or active human supervision. Fig.\ref{fig:filtering} shows a visualization of the human datasets after filtering the noise, where the processing of the initial cluttered trajectories leaves well-organized fragments of trajectories pointing towards the target.

\section{Experiments}
In this section, we experimentally answer the following questions. 1) Is there an optimal range of noise angle? 2) Does evaluating the data with fine-grained confidence scores improve the algorithm performance? 3) Which is more aligned with human intent, penalizing noise or simply filtering it?
\subsection{Experiments Setting}
We develop a simulated robotic teleoperation system that leverages the Omega.3 haptic device's input signaling to gather human demonstrations. This system integrates smoothly with Robosuite, which equips the IIWA robotic arm with an OSC-POSE controller. The 
action inputs $a$ consist of position vectors $\{x, y, z\}$, direction vectors $\{row, pitch, yaw\}$, and a grasping signal $\{grasp\}$. The structure of the human datasets is set as $\mathcal{D}_E=\{\tau_1,\tau_2,...,\tau_n\}$, where $\tau_n$ is the trajectory which stores each transition $(s_t,a_t,s_{t+1})$ according to the decision-making sequence, such as $\tau_n=\{(s_0,a_0,s_1),..., (s_t,a_t,s_{t+1})\}$ with $t=T_n$. Here, $T_n$ is the trajectory's length that refers to the number of decision steps of the trajectory. After familiarizing the demonstrators with our system, we limited them to complete each demonstration within 25s (trajectory length $T_n$ is set to 500 steps). As shown in Fig.\ref{fig:Tele}, using trajectory's length as a criterion, we categorized the human datasets into three types: Better, Worse, and Failed, each containing 30 trajectories. Among them, the Better datasets refers to trajectories that completed the task quickly, and the Failed datasets refers to trajectories that did not complete the task within the required time. The learning rate of Actor and Critic networks are set to 5e-6 and the total number of training steps is 5e5.
\begin{figure}[t]
	\centering
	\includegraphics[width=0.48\textwidth]{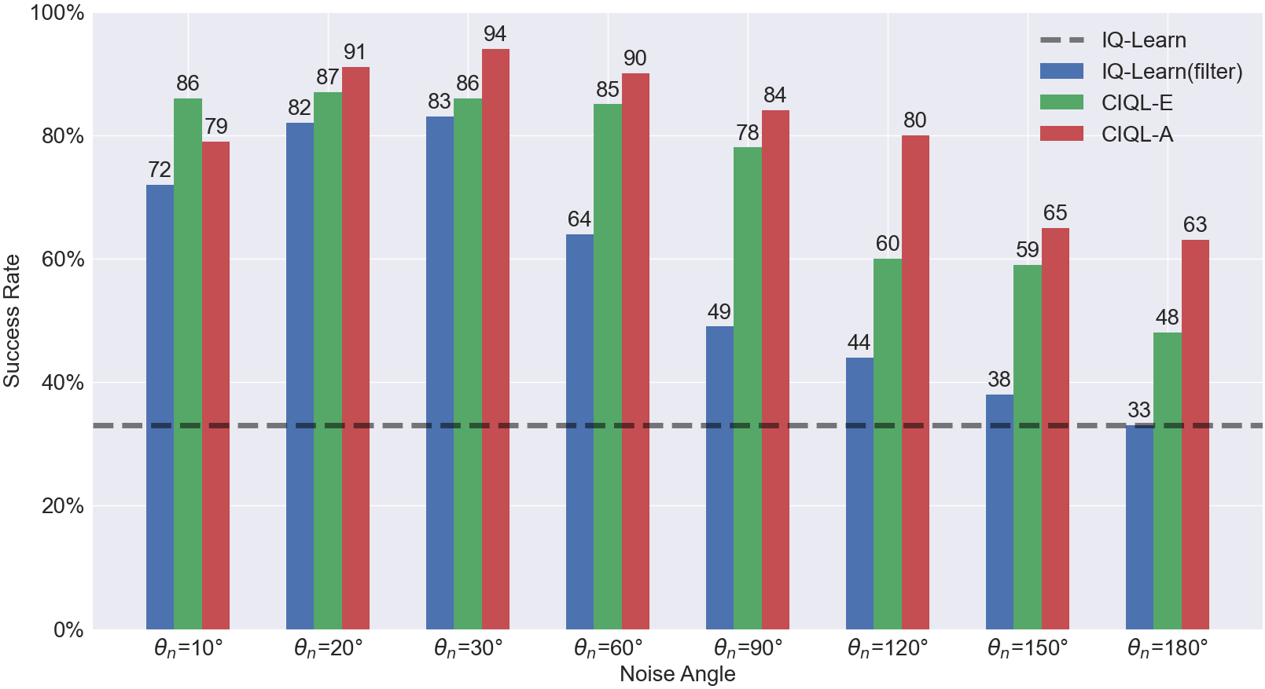}
	\caption{\textbf{Effect of Noise Angle.} IQ-Learn: Baseline algorithm; IQ-Learn (filter): Filtering noise without using confidence, it becomes IQ-Learn when $\theta_n$ is set to 180°. CIQL-E: Filtering noise and using confidence; CIQL-A: Penalizing noise and using confidence.}
	\label{BoundaryAngle}
\end{figure}

\subsection{Different Noise Angles and Datasets}
In order to compare whether a fine-grained confidence evaluation for the data improves the performance of the baseline algorithm, we introduce another algorithm for ablation studies, IQ-Learn(filter), which adopts our method to filter the noise without using confidence. We perform ablation experiments using a combination of Better-Worse-Failed(90) (30 trajectories per datasets) and explore the effect of noise angle on the success rate from 10$^{\circ}$ to 180$^{\circ}$, as shown in Fig.\ref{BoundaryAngle}. The experimental results show that the performance has an inverted U-shaped relationship with the noise angle: the performance increases when the noise angle is below a certain threshold, but decreases when it exceeds the threshold. The threshold is the optimal noise angle for the linear task, which is between 20$^{\circ}$ and 60$^{\circ}$. This phenomenon can be explained by the role of thresholds in dividing the data. Data above this threshold are considered to be true noise and can adversely affect performance. Reducing the noise angle may inadvertently filter out data that has the potential to improve performance while reducing the diversity of the data.

\begin{figure}[t]
	\centering
	\includegraphics[width=0.48\textwidth]{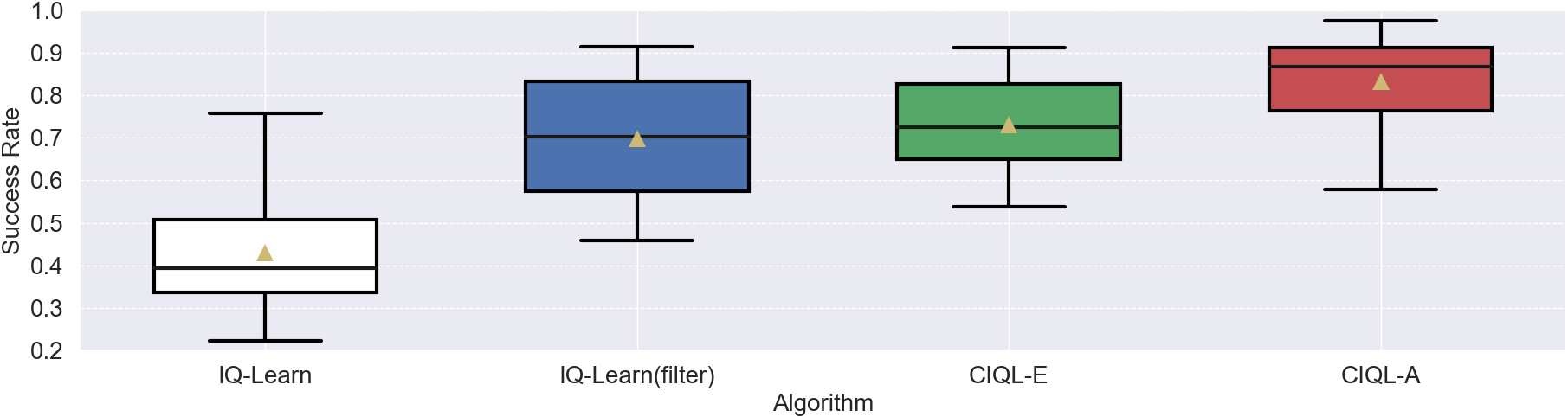}
	\caption{\textbf{Algorithm Performance.} We evaluate the policy model using the average success rate of 10 random seeds, each testing 100 trajectories (1000 trajectories in total). We use all combinations of the datasets in TABLE \ref{table1} and the yellow triangle represents the average success rate.}
	\label{box}
\end{figure}

We test the performance of each algorithm for different combinations of datasets with the noise angle set to 60$^{\circ}$, as shown in TABLE \ref{table1} and Fig.\ref{box}. Compared to the baseline algorithm, these algorithms are ranked in terms of average success across all combinations of datasets: CIQL-A(40.3$\%$) $>$ CIQL-E(30.1$\%$) $>$ CIQL(filter, 26.8$\%$) $>$ IQ-Learn. After evaluating the data using our method on all noise angles and datasets, the baseline algorithm shows a significant improvement in performance in terms of both mean and variance. This proves the robustness and effectiveness of our method. In-depth analysis shows that CIQL-A improves the success rate by 10.2$\%$ over CIQL-E in tests on all datasets. This proves that penalizing noise is better than filtering noise when confidence is also used. Comparing CIQL-E with IQ-Learn(filter) shows that evaluating the remaining data with fine-grained confidence after filtering the noise can still improve the baseline algorithm's performance. The comparison of IQ-Learn(filter) with IQ-Learn indicates that the noise has a very serious impact on the baseline algorithm.

\begin{table}[t]
	%\caption{}
	%\textbf{Table 1}~~\textbf{Effect of Dataset.} 
	\caption{Average Success Rate for Different Combinations of Datasets}
	\begin{center}
		\setlength{\tabcolsep}{0.8mm}{
			\begin{tabular}{ccccc}
				\toprule
				Datasets&IQ-Learn&IQ-Learn(filter)&CIQL-E&CIQL-A\\
				\midrule
				Better(30)&66.9$\pm$4.7 &86.4$\pm$3.3 &84.4$\pm$3.4 &\textbf{86.9$\pm$4.0} \\
				Worse(30)&38.1$\pm$4.3 &51.5$\pm$3.9 &61.8$\pm$5.9 &\textbf{71.9$\pm$3.5} \\
				\midrule
				Better-Worse(30)&27.7$\pm$5.4 &51.4$\pm$4.0 &60.7$\pm$5.2 &\textbf{91.9$\pm$1.4} \\
				Better-Failed(30)&57.0$\pm$5.7 &78.3$\pm$4.2 &82.2$\pm$4.2 &\textbf{84.1$\pm$3.4} \\
				Worse-Failed(30)&32.4$\pm$5.4 &62.5$\pm$4.3 &\textbf{67.6$\pm$4.5} &65.9$\pm$5.5 \\
				\midrule
				Better-Worse-Failed(30)&47.3$\pm$4.6 &76.0$\pm$6.2 &71.7$\pm$4.8&\textbf{81.0$\pm$4.2} \\
				Better-Worse-Failed(60)&41.3$\pm$5.3 &87.8$\pm$3.0 &70.9$\pm$3.3&\textbf{94.0$\pm$2.2} \\
				Better-Worse-Failed(90)&32.9$\pm$4.9 &64.2$\pm$4.6 &85.0$\pm$2.8&\textbf{90.3$\pm$2.1} \\
				\bottomrule
		\end{tabular}}
	\end{center}
	\label{table1}
\end{table}

\subsection{Alignment Testing}
\begin{figure}[t]
	\centering
	\subfloat[CIQL-A’s recovered reward]{\includegraphics[width=0.22\textwidth]{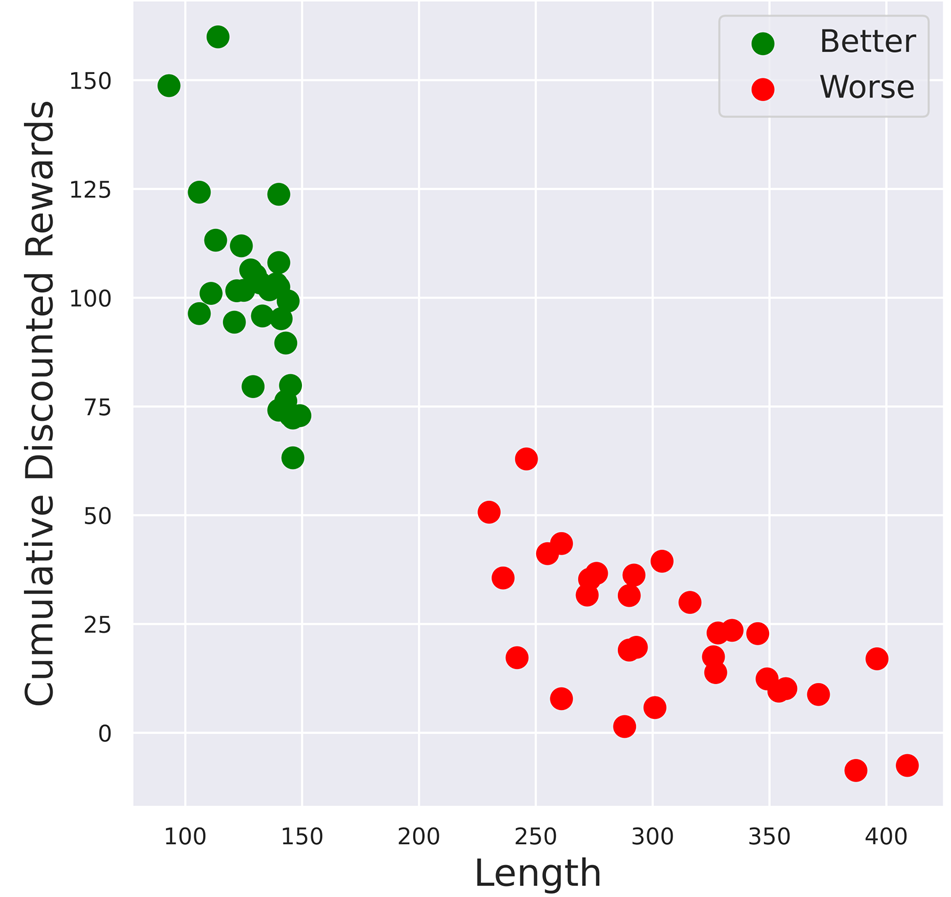}\label{fig: CIQLA-reward}}
	\subfloat[CIQL-E’s recovered reward]{\includegraphics[width=0.22\textwidth]{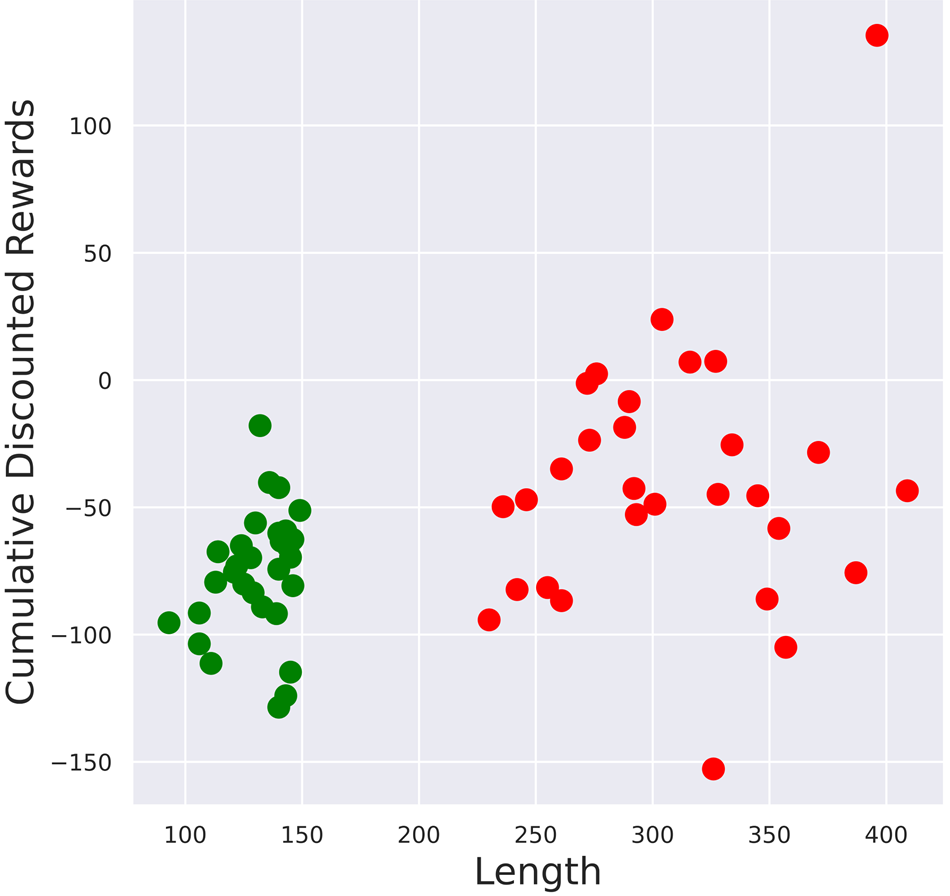}\label{fig: CIQLE-reward}}
	\caption{\textbf{Reward Alignment Testing.} Use the recovered reward function learned in the Better-Worse-Failed datasets (60, $\theta$=60$^\circ$) to calculate the cumulative discounted reward for each trajectory in demonstrations. The Pearson correlation for CIQL-A is -0.92 and for CIQL-E is 0.46.}
	\label{IRLward}
\end{figure}

The trajectory's length can be used as a metric of human performance, the shorter the length, the better the performance tends to be\cite{mandlekar2021matters}. We conduct alignment testing from the perspectives of reward and policy. From the reward perspective, we use the recovered reward function to compute the cumulative discounted rewards $\eta_{\tau_i}=\sum_{t=0}^{T_i}\gamma^tr(s_t,a_t)$ for each trajectory, as shown in Fig.\ref{IRLward}. The results indicate that under the reward function evaluation recovered by CIQL-A, the trajectory's length is highly negatively correlated with its cumulative discounted rewards, with a Pearson correlation of -0.92. In contrast, for CIQL-E, the correlation is 0.46. Since noise does not contribute to the learned reward of CIQL-E, causing the reward function failing to evaluate it correctly, whereas the reward function of CIQL-A gives a negative value to this noise, making the reward function more aligned with human intent. From the policy perspective, we directly compare the performance of the four algorithms in accomplishing the task, and specific simulations can be found in this \href{https://github.com/XizoB/CIQL}{VIDEO}. In one hundred independent experiments, the total time for each algorithm to complete the task is as follows: IQ-Learn(346s), IQ-Learn(filter, 290s), CIQL-E(255s) and CIQL-A(178s), which can be seen that CIQL-A performs more perfectly. Additionally, the demonstrators are instructed to avoid touching the table during the demonstration, while the robot seems not to have learned this feature. CIQL-E touched the table 17 times in the experiment, while CIQL-A touched it only 8 times, which is more aligned with the human intent.
\begin{figure}[t]
	\centering
	\includegraphics[width=0.46\textwidth]{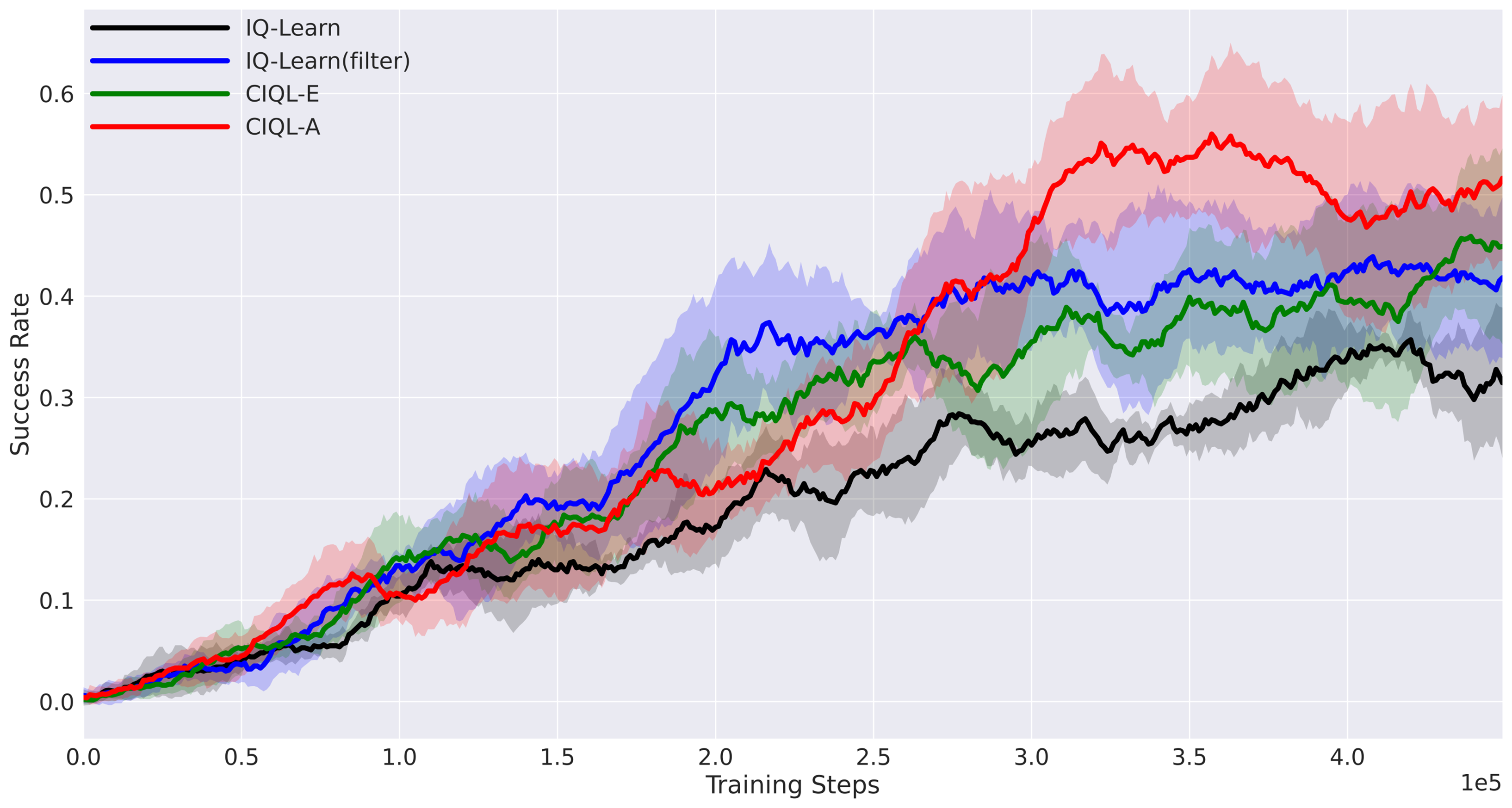}
	\caption{\textbf{Grasping Success Rate in Block Stacking.} Each algorithm is trained using five random seeds. We select the moving average success rate of every ten trajectories as the criterion for choosing the optimal model in training.}
	\label{successrate}
\end{figure}
\begin{table}[t]
	%\caption{}
	\caption{Average  Success Rate In Block Stacking}
	%\textbf{Table 1}~~\textbf{Effect of Dataset.} 
	\begin{center}
		\setlength{\tabcolsep}{0.8mm}{
			\begin{tabular}{ccccc}
				\toprule
				Block Stacking &IQ-Learn&IQ-Learn(filter)&CIQL-E&CIQL-A\\
				\midrule
			    Linear Grasping & 32.9$\pm$3.7 &52.8$\pm$4.5 &57.8$\pm$4.4 &\textbf{66.2$\pm$3.0} \\
				Total &1.5$\pm$1.0 &3.2$\pm$1.6 &6.2$\pm$1.7 &\textbf{7.5$\pm$2.6} \\
				\bottomrule
		\end{tabular}}
	\end{center}
	\label{table2}
\end{table}

\subsection{Multi-stage Task Testing}
To generalize our framework to multi-stage tasks, we additionally collect the Better-Worse(60) datasets for a block stacking experiment and set it up as a long sequential decision-making task. The number of decision steps in our datasets ranges from 100 to 500, and this mixed-policy datasets is challenging for imitation learning. The demonstrators required placing a red wooden block with a side length of 4cm on top of a green wooden block with a side length of 5cm. This includes two actuator actions of grasping and placing, as well as robot movement stage such as linear approach, lifting, rotation, and alignment. Our method is only used for the linear grasping task in block stacking, evaluating the imperfect data when approaching the target. We set the noise angle to 60$^\circ$, the confidence of the data points for grasping and placing to 1.5, and the confidence of all other data points to 1. Combining the results in Fig.\ref{successrate} and TABLE \ref{table2}, it can be found that our method can greatly enhance the success rate of grasping by almost 33.3$\%$, which leads to a 6$\%$ increase in the success rate of the overall task. This also validates our claim that any task with linear stage can benefit from the enhancement of our method on the performance of the baseline algorithm. More importantly, we only require human demonstrations, without the need for additional information or strict assumptions. 

\begin{figure}[t]
	\centering
	\includegraphics[width=0.42\textwidth]{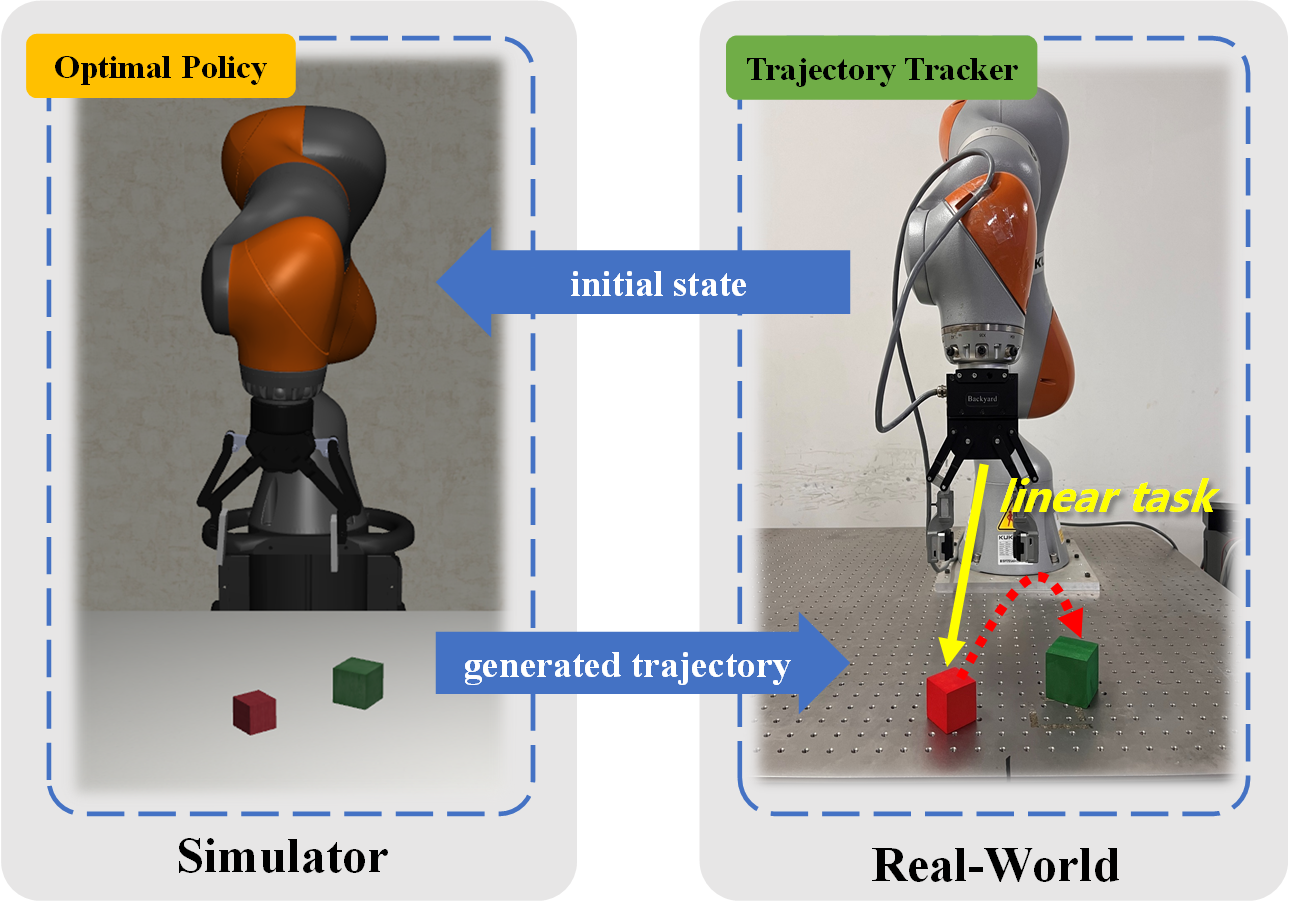}
	\caption{\textbf{Sim2Real.} Transfer the real-world state as an initial state to the simulator and use the PD controller to track the trajectory generated by the optimal policy. Our method only works for the linear stage in stacking block, i.e., evaluating imperfect data when approaching the grasping target.}
	\label{Sim2Real}
\end{figure}

To strengthen the practicality of the research, we conduct a Sim2Real experiment using the CIQL-A algorithm, which can provide a stable nearly straight trajectory for the robot to grasp the target, as shown in Fig.\ref{Sim2Real}. In the real world, we do not need to generate many points for a trajectory, which may increase the robot's probability of making mistakes and accumulate errors. Actually, the robot requires only four or five decision steps to complete the block stacking. In the future, we will explore the optimal decision step setting intervals for different stages of the robot's task, which will provide a reference for the real-world application of the robot.

\subsection{Discussion}
In different noise angle experiments, we find that when the noise angle is set to a small value, especially to 10$^{\circ}$, some optimal data may be incorrectly recognized as noise, which in turn negatively affects the penalizing noise algorithm, CIQL-A, as shown in the Fig.\ref{BoundaryAngle}. CIQL-E can be severely affected by the type of datasets and is sometimes inferior to method that filter noise directly. Furthermore, in the datasets combination of Better-Worse-Failed, we find that the algorithm's performance does not improve with the increase in the amount of data. These can be ascribed to the failure to find the optimal noise angle or the correct confidence function, and incorrect confidence scores can adversely affect the algorithm.

\section{Conclusion}
\textbf{Summary.} We develop a general framework for using confidence scores based on the IQ-Learn algorithm and provide two optimal policy learning methods. We introduce the concept of noise angle and propose a transition-based evaluation method to provide fine-grained confidence scores for datasets without additional information. Through ablation experiments, we find that the optimal interval of the noise angle for linear task is between 20$^\circ$ and 60$^\circ$. Combining experiments with different combinations of datasets, we find that our method improves the performance of the baseline algorithm by 40.3$\%$. Furthermore, we find that penalizing noise is more effective and more aligned with human intent than simply filtering it.

\textbf{Limitation and Future Work.} 
While our data confidence evaluation method can improve the baseline algorithm's performance, they are sensitive to the choice of noise angle, and an incorrect confidence score can instead adversely affect the baseline algorithm. In order to accomplish complex tasks, it is sometimes necessary to avoid obstacles while approaching a target, and it is a challenge to set the noise angle for this scenario. The robotics tasks are very sensitive to keypoints, which is critical to the ability of imitation learning, especially inverse reinforcement learning, to achieve good results, and we will consider how to address this issue in the future. In addition, supervised learning tasks usually select the model with the least loss, while robotics tasks focus more on the success rate under safe conditions. Therefore, evaluating whether a policy is aligned with human intents is an extremely important and challenging work.

%%%%%%%%%%%%%%%%%%%%%%%%%%%%%%%%%%%%%%%%%%%%%%%%%%%%%%%%%%%%%%%%%%%%%%%%%%%%%%%%
%\begin{thebibliography}{99}
%\end{thebibliography}

\bibliographystyle{IEEEtran}
\bibliography{ref}

\vfill
%%%%%%%%%%%%%%%%%%%%%%%%%%%%%%%%%%%%%%%%%%%%%%%%%%%%%%%%%%%%%%%%%%%%%%%%%%%%%%%%

\end{document}